\documentclass[10pt,twocolumn,letterpaper]{article}

\usepackage{iccv}
\usepackage{times}
\usepackage{epsfig}
\usepackage{graphicx}
\usepackage{amsmath}
\usepackage{amssymb}
\usepackage{appendix}
\usepackage{algorithm}
\usepackage{algpseudocode}

\newcommand{\bi}{\begin{itemize}}
\newcommand{\ei}{\end{itemize}}
\newcommand{\bal}{\begin{align}}
\newcommand{\eal}{\end{align}}

\newcommand{\EE}{\mathbb{E}}
\newcommand{\PP}{\mathbb{P}}

\newcommand{\bx}{\mathbf{x}}
\newcommand{\by}{\mathbf{y}}

\newcommand{\bI}{\mathbf{I}}

\newcommand{\bz}{\mathbf{z}}

\newcommand{\bG}{\mathbf{G}}

\newcommand{\cN}{\mathcal{N}}

\newcommand{\cP}{\mathcal{P}}

\newcommand{\cG}{\mathcal{G}}
\newcommand{\cF}{\mathcal{F}}

\newcommand{\eps}{\epsilon}

\def\<{\langle}
\def\>{\rangle}

\newtheorem{theorem}{\textbf{Theorem}}
\newtheorem{lemma}{\textbf{Lemma}}

\usepackage[pagebackref=true,breaklinks=true,letterpaper=true,colorlinks,bookmarks=false]{hyperref}

\iccvfinalcopy 

\def\iccvPaperID{4881} 

\ificcvfinal\fi
\begin{document}

\title{Normalized Wasserstein for Mixture Distributions with Applications in Adversarial Learning and Domain Adaptation}

\author{Yogesh Balaji\\
Department of Computer Science\\
University of Maryland \\
{\tt\small yogesh@cs.umd.edu}
\and
Rama Chellappa \\
UMIACS \\
University of Maryland\\
{\tt\small rama@umiacs.umd.edu}
\and
Soheil Feizi \\
Department of Computer Science\\
University of Maryland \\
{\tt\small sfeizi@cs.umd.edu}
}
\maketitle

\begin{abstract}
Understanding proper distance measures between distributions is at the core of several learning tasks such as generative models, domain adaptation, clustering, etc. In this work, we focus on {\it mixture distributions} that arise naturally in several application domains where the data contains different sub-populations. For mixture distributions, established distance measures such as the Wasserstein distance do not take into account imbalanced mixture proportions. Thus, even if two mixture distributions have identical mixture components but different mixture proportions, the Wasserstein distance between them will be large. This often leads to undesired results in distance-based learning methods for mixture distributions. In this paper, we resolve this issue by introducing the {\it Normalized Wasserstein} measure. The key idea is to introduce mixture proportions as optimization variables, effectively normalizing  mixture proportions in the Wasserstein formulation. Using the proposed normalized Wasserstein measure leads to significant performance gains for mixture distributions with imbalanced mixture proportions compared to the vanilla Wasserstein distance. We demonstrate the effectiveness of the proposed measure in GANs, domain adaptation and adversarial clustering in several benchmark datasets.
\end{abstract}

\section{Introduction}
Quantifying distances between probability distributions is a fundamental problem in machine learning and statistics with several applications in generative models, domain adaptation, clustering, etc. Popular probability distance measures include {\it optimal transport} measures such as the Wasserstein distance \cite{villani2008optimal} and {\it divergence} measures such as the Kullback-Leibler (KL) divergence \cite{cover2012elements}.  

Classical distance measures, however, can lead to some issues for mixture distributions. A {\it mixture distribution} is the probability distribution of a random variable $X$ where $X=X_i$ with probability $\pi_i$ for $1\leq i\leq k$. $k$ is the number of mixture components and $\pi=[\pi_1,...,\pi_k]^T$ is the vector of {\it mixture (or mode) proportions}. The probability distribution of each $X_i$ is referred to as a {\it mixture component} (or, a mode). Mixture distributions arise naturally in different applications where the data contains two or more sub-populations. For example, image datasets with different labels can be viewed as a mixture (or, multi-modal) distribution where samples with the same label characterize a specific mixture component. 

If two mixture distributions have exactly same mixture components (i.e. same $X_i$'s) with different mixture proportions (i.e. different $\pi$'s), classical distance measures between the two will be large. This can lead to undesired results in several distance-based machine learning methods. To illustrate this issue, consider the {\it Wasserstein distance} between two distributions $\PP_X$ and $\PP_{Y}$, defined as \cite{villani2008optimal}
\begin{align}\label{eq:W}
&W(\PP_X,\PP_{Y}):=\min_{\PP_{X,Y}}~ \EE\left[\|X-Y\|\right],\\
&\text{marginal}_X(\PP_{X,Y})=\PP_X, \text{  marginal}_{Y}(\PP_{X,Y})=\PP_{Y}\nonumber
\end{align}
where $\PP_{X,Y}$ is the joint distribution (or coupling) whose marginal distributions are equal to $\PP_X$ and $\PP_{Y}$. When no confusion arises and to simplify notation, in some equations, we use $W(X,Y)$ notation instead of $W(\PP_X,\PP_{Y})$. 

\begin{figure*}[t]
\centering
\includegraphics[width=\textwidth]{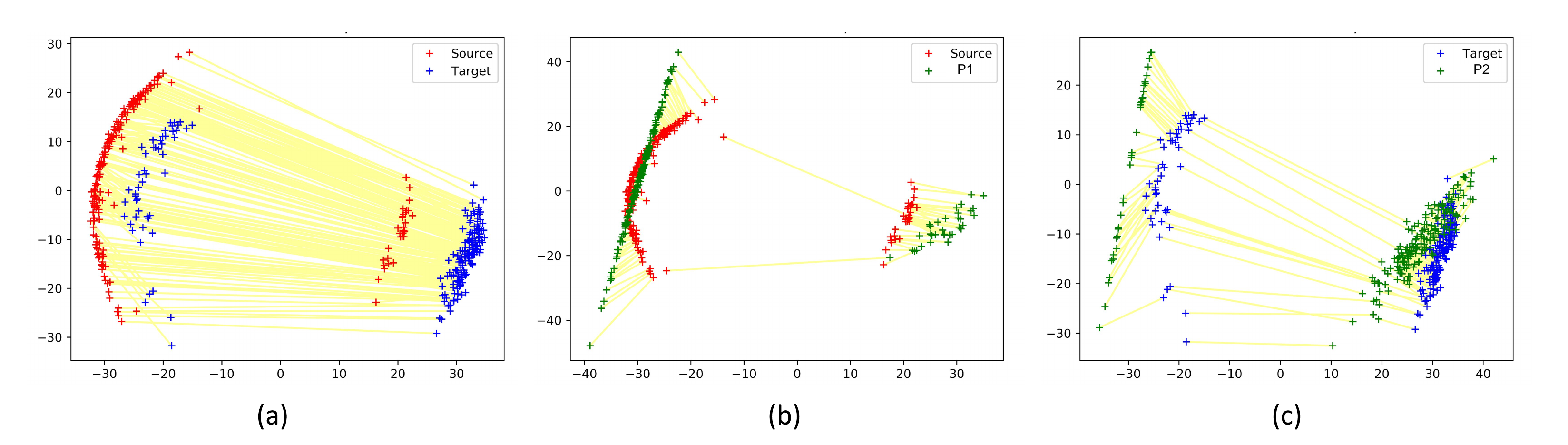}
\caption{An illustration of the effectiveness of the proposed Normalized Wasserstein measure in domain adaptation. The source domain (shown in red) and the target domain (shown in blue) have two modes with different mode proportions. (a) The couplings computed by estimating Wasserstein distance between source and target distributions (shown in yellow lines) match several samples from incorrect and distant mode components. (b,c) Our proposed normalized Wasserstein measure \eqref{eq:N-W} constructs intermediate mixture distributions $\PP_1$ and $\PP_2$ (shown in green) with similar mixture components to source and target distributions, respectively, but with optimized mixture proportions. This significantly reduces the number of couplings between samples from incorrect modes and leads to $42 \%$ decrease in target loss in domain adaptation compared to the baseline.}
\label{fig:title_fig} 
\end{figure*}

The Wasserstein distance optimization is over all joint distributions (couplings) $\PP_{X,Y}$ whose marginal distributions {\it match} exactly with input distributions $\PP_X$ and $\PP_Y$. This requirement can cause issues when $\PP_X$ and $\PP_Y$ are mixture distributions with different mixture proportions. In this case, due to the marginal constraints, samples belonging to very different mixture components will have to be coupled together  in $\PP_{X,Y}$ (e.g. Figure~\ref{fig:title_fig}(a)). Thus, using this distance measure can then lead to undesirable outcomes in problems such as domain adaptation. This motivates the need for developing a new distance measure to take into account mode imbalances in mixture distributions.

In this paper, we propose a new distance measure that resolves the issue of imbalanced mixture proportions for multi-modal distributions. Our developments  focus on a class of optimal transport measures, namely the Wasserstein distance Eq~\eqref{eq:W}. However, our ideas can be extended naturally to other distance measures (eg. adversarial distances~\cite{ganin2015DANN}) as well.

Let $\bG$ be an array of generator functions with $k$ components defined as $\bG:=[\bG_1,...,\bG_k]$. Let $\PP_{\bG,\pi}$ be a mixture probability distribution for a random variable $X$ where $X=\bG_i(Z)$ with probability $\pi_i$ for $1\leq i\leq k$. Throughout the paper, we assume that $Z$ has a normal distribution. 

By relaxing the marginal constraints of the classical Wasserstein distance \eqref{eq:W}, we introduce the {\it Normalized Wasserstein measure} ({\it NW measure}) as follows: 
\begin{align*}
&W_N(\PP_{X},\PP_{Y})\\
&:=\min_{\bG,\pi^{(1)},\pi^{(2)}} ~W(\PP_{X},\PP_{\bG,\pi^{(1)}})+W(\PP_{Y},\PP_{\bG,\pi^{(2)}}).\nonumber
\end{align*}
There are two key ideas in this definition that help resolve mode imbalance issues for mixture distributions. First, instead of directly measuring the Wasserstein distance between $\PP_X$ and $\PP_Y$, we construct two intermediate (and potentially mixture) distributions, namely $\PP_{\bG,\pi^{(1)}}$ and $\PP_{\bG,\pi^{(2)}}$. These two distributions have the same mixture components (i.e. same $\bG$) but can have different mixture proportions (i.e. $\pi^{(1)}$ and $\pi^{(2)}$ can be different). Second, mixture proportions, $\pi^{(1)}$ and $\pi^{(2)}$, are considered as optimization variables. This effectively {\it normalizes} mixture proportions before Wasserstein distance computations. See an example in Figure~\ref{fig:title_fig} (b, c) for a visualization of $\PP_{\bG,\pi^{(1)}}$ and $\PP_{\bG,\pi^{(2)}}$, and the re-normalization step.

In this paper, we show the effectiveness of the proposed Normalized Wasserstein measure in three application domains. In each case, the performance of our proposed method significantly improves against baselines when input datasets are mixture distributions with imbalanced mixture proportions. Below, we briefly highlight these results:

{\bfseries Domain Adaptation:} In Section \ref{sec:domain-adapt}, we formulate the problem of domain adaptation as minimizing the normalized Wasserstein measure between source and target feature distributions. On classification tasks with imbalanced datasets, our method significantly outperforms baselines (e.g. $\sim 20\%$ gain in synthetic to real adaptation on VISDA-3 dataset).

{\bfseries GANs:} In Section \ref{sec:gans}, we use the normalized Wasserstein measure in GAN's formulation to train mixture models with varying mode proportions. We show that such a generative model can help capture rare modes, decrease the complexity of the generator, and re-normalize an imbalanced dataset.

{\bfseries Adversarial Clustering:} In Section \ref{sec:adv_clustering}, we formulate the clustering problem as an adversarial learning task using Normalized Wasserstein measure.

\section{Normalized Wasserstein Measure}\label{sec:N-W}
In this section, we introduce the {\it normalized Wasserstein} measure and discuss its properties. Recall that $\bG$ is an array of generator functions defined as $\bG:=[\bG_1,...,\bG_k]$ where $\bG_i: \mathbb{R}^r\to \mathbb{R}^d$. Let $\cG$ be the set of all possible $\bG$ function arrays. Let $\mathbf{\pi}$ be a discrete probability mass function with $k$ elements, i.e. $\pi=[\pi_1,\pi_2,\cdots,\pi_k]$ where $\pi_i \geq 0$ and $\sum_i \pi_i=1$. Let $\Pi$ be the set of all possible $\pi$'s. 

Let $\PP_{\bG,\pi}$ be a  mixture distribution, i.e. it is the probability distribution of a random variable $X$ such that $X=\bG_i(Z)$ with probability $\pi_i$ for $1\leq i\leq k$. We assume that $Z$ has a normal density, i.e. $Z\sim \cN(\mathbf{0},\bI)$. We refer to $\bG$ and $\pi$ as mixture components and proportions, respectively. The set of all such mixture distributions is defined as:
\begin{align}\label{def:order-one-mixture}
\cP_{\bG,k}:=\left\{\PP_{\bG,\pi}: \bG\in \cG, \pi \in \Pi \right\}
\end{align}
where $k$ is the number of mixture components.
Given two distributions $\PP_X$ and $\PP_Y$ belonging to the family of mixture distributions $\cP_{\bG,k}$, we are interested in defining a distance measure agnostic to differences in mode proportions, but sensitive to shifts in mode components, i.e., the distance function should have high values only when mode components of $\PP_X$ and $\PP_Y$ differ. If $\PP_X$ and $\PP_Y$ have the same mode components but differ only in mode proportions, the distance should be low.

The main idea is to introduce mixture proportions as optimization variables in the Wasserstein distance formulation \eqref{eq:W}. This leads to the following distance measure which we refer to as the {\it Normalized Wasserstein measure} (NW measure), $W_N(\PP_{X},\PP_{Y})$, defined as:
\begin{align}\label{eq:N-W}
\min_{\bG,\pi^{(1)},\pi^{(2)}} ~ &W(\PP_{X},\PP_{\bG,\pi^{(1)}})+W(\PP_{Y},\PP_{\bG,\pi^{(2)}})\\
&\sum_{j=1}^{k} \pi^{(i)}_j =1 \quad ~i=1,2,\nonumber\\
& \pi^{(i)}_j \geq 0 \quad \quad\quad  1\leq j\leq k,~~ i=1,2.\nonumber
\end{align}
Since the normalized Wasserstein's optimization \eqref{eq:N-W} includes mixture proportions $\pi^{(1)}$ and $\pi^{(2)}$ as optimization variables, if two mixture distributions have similar mixture components with different mixture proportions (i.e. $\PP_X=\PP_{\bG,\pi^{(1)}}$ and $\PP_{Y}=\PP_{\bG,\pi^{(2)}}$), although the Wasserstein distance between the two can be large, the introduced normalized Wasserstein measure between the two will be zero. Note that $W_N$ is defined with respect to a set of generator functions $\bG=[\bG_1,...,\bG_k]$. However, to simplify the notation, we make this dependency implicit. We would like to point our that our proposed NW measure is a \emph{semi-distance} measure (and not a distance) since it does not satisfy all properties of a distance measure. Please refer to Appendix for more details.

To compute the NW measure, we use an alternating gradient descent approach similar to the dual computation of the Wasserstein distance \cite{arjovsky2017wasserstein}. Moreover, we impose the $\pi$ constraints using a soft-max function. Please refer to Appendix.~\ref{app:duality} for more details.

To illustrate how NW measure is agnostic to mode imbalances between distributions , consider an unsupervised {\it domain adaptation} problem with MNIST-2 (i.e. a dataset with two classes: digits 1 and 2 from MNIST) as the source dataset, and noisy MNIST-2 (i.e. a noisy version of it) as the target dataset (details of this example is presented in Section~\ref{sec:DA_unsup}). The source dataset has $4/5$ digits one and $1/5$ digits two, while the target dataset has $1/5$ noisy digits one and $4/5$ noisy digits two. The couplings produced by estimating the Wasserstein distance between the two distributions is shown in yellow lines in Figure~\ref{fig:title_fig}-a. We observe that there are many couplings between samples from incorrect mixture components. The normalized Wasserstein measure, on the other hand, constructs intermediate mode-normalized distributions $\PP_1$ and $\PP_2$, which get coupled to the correct modes of source and target distributions, respectively (see panels (b) and (c) in Figure \ref{fig:title_fig})).

\section{Theoretical Results}\label{sec:theory}
For NW measure to work effectively, the number of modes $k$ in NW formulation (Eq.~\eqref{eq:N-W}) must be chosen appropriately. For instance, given two mixture distributions with $k$ components each, Normalized Wasserstein measure with $2k$ modes would always give $0$ value. In this section, we provide some theoretical conditions under which the number of modes can be estimated accurately. We begin by making the following assumptions for two mixture distributions $X$ and $Y$ whose NW distance we wish to compute:
\begin{itemize}
    \item (A1) If mode $i$ in distribution $X$ and mode $j$ in distribution $Y$ belong to the same mixture component, then their Wasserstein distance is $\leq \epsilon$ i.e., if $X_i$ and $Y_j$ correspond to the same component, $W(\PP_{X_i}, \PP_{Y_j}) < \epsilon$.
    \item (A2) The minimum Wasserstein distance between any two modes of one mixture distribution is at least $\delta$ i.e., $W(\PP_{X_i}, \PP_{X_j}) > \delta$ and $W(\PP_{Y_i}, \PP_{Y_j}) > \delta$ $\forall i \neq j$. Also, non-overlapping modes between $X$ and $Y$ are separated by $\delta$ i.e., for non-overlapping modes $X_i$ and $Y_j$,  $W(\PP_{X_i}, \PP_{Y_j}) > \delta$.  This ensures that modes are well-separated.
    \item (A3) We assume that each mode $X_i$ and $Y_i$ have density at least $\eta$ i.e., $\PP_{X_i} \geq \eta \quad \forall i$, $\PP_{Y_i} \geq \eta \quad \forall i$. This ensures that every mode proportion is at least $\eta$.
    \item (A4) Each generator $\bG_i$ is powerful enough to capture exactly one mode of distribution $\PP_X$ or $\PP_Y$.
\end{itemize}
\begin{theorem}\label{thm:main}
Let $\mathbb{P}_X$ and $\mathbb{P}_Y$ be two mixture distributions satisfying (A1)-(A4) with $n_{1}$ and $n_{2}$ mixture components, respectively, where $r$ of them are overlapping. Let $k^*=n_1 + n_2 - r$. Then, $k^*$ is smallest $k$ for which $NW(k)$ is small ($O(\epsilon)$) and $NW(k) - NW(k-1)$ is relatively large (in the $O(\delta \eta)$ )
\end{theorem}
The proof is presented in Appendix.~\ref{app:proof}. All assumptions made are reasonable and hold in most practical situations: (A1)-(A3) enforces that non-overlapping modes in mixture distribitions are separated, and overlapping modes are close in Wasserstein distance. To enforce (A4), we need to prevent multi-mode generation in one mode of $\bG$. This can be satisfied by using the regularizer in Eq.~\eqref{eq:reg}. Note that in the above theorem, $k^*$ is the optimal $k$ that should be used in the Normalized Wasserstein formulation. The theorem presents a way to estimate $k^*$. Please refer to Section~\ref{sec:choosing_num_modes} for experimental results.  In many applications like domain adaption, however, the number of components $k$ is known beforehand, and this step can be skipped.

\section{Normalized Wasserstein for Domain Adaptation under covariate and label shift}\label{sec:domain-adapt} 
In this section, we demonstrate the effectiveness of the NW measure in Unsupervised Domain Adaptation (UDA) both for supervised (e.g. classification) and unsupervised (e.g. denoising) tasks. Note that the term \emph{unsupervised} in UDA means that the label information in the target domain is unknown while \emph{unsupervised} tasks mean that the label information in the source domain is unknown.  

First, we consider domain adaptation for a classification task. Let $(X_s,Y_s)$ represent the source domain while $(X_t,Y_t)$ denote the target domain. Since we deal with the classification setup, we have $Y_s,Y_t \in \{1,2,...,k \}$. A common formulation for the domain adaptation problem is to transform $X_s$ and $X_t$ to a feature space where the {\it distance} between the source and target feature distributions is sufficiently small, while a good classifier can be computed for the source domain in that space~\cite{ganin2015DANN}. In this case, one solves the following optimization:
\begin{align}\label{opt:domain-supervised}
\min_{f\in \cF}~~  L_{cl}\left(f(X_s),Y_s\right) + \lambda\text{ dist}\left(f(X_s),f(X_t)\right)
\end{align}
where $\lambda$ is an adaptation parameter and $L_{cl}$ is the empirical classification loss function (e.g. the cross-entropy loss). The distance function between distributions can be adversarial distances~\cite{ganin2015DANN,tzeng2017adversarial}, the Wasserstein distance~\cite{shen2017WDDA}, or MMD-based distances~\cite{long2015DAN, long2016RTN}.

When $X_s$ and $X_t$ are mixture distributions (which is often the case as each label corresponds to one mixture component) with different mixture proportions, the use of these classical distance measures can lead to the computation of inappropriate transformation and classification functions. In this case, we propose to use the NW measure as the distance function. Computing the NW measure requires training mixture components $\bG$ and mode proportions $\pi^{(1)}, \pi^{(2)}$. To simplify the computation, we make use of the fact that labels for the source domain (i.e. $Y_s$) are known, thus source mixture components can be identified using these labels. Using this information, we can avoid the need for computing $\bG$ directly and use the conditional source feature distributions as a proxy for the mixture components as follows:
\begin{align}\label{eq:matching-dist}
&\bG_i(Z) \stackrel{\text{dist}}{=} f(X^{(i)}_s), \\
&X^{(i)}_s = \{ X_s| Y_s=i\}, \quad \forall 1\leq i\leq k,\nonumber
\end{align}
where $\stackrel{\text{dist}}{=}$ means matching distributions. Using \eqref{eq:matching-dist}, the formulation for domain adaptation can be written as
\begin{align}\label{opt:domain-supervised-simplified}
\min_{f\in \cF} ~ \min_{\pi} ~ & L_{cl}\left(X_{s},Y_{s}\right) + \lambda W\left( \sum_i \pi^{(i)}f(X_s^{(i)})  ,f(X_t)\right) .
\end{align}
The above formulation can be seen as a version of instance weighting as source samples in $X_s^{(i)}$ are weighted by $\pi_i$. Instance weighting mechanisms have been well studied for domain adaptation~\cite{yan2017mind_classbias,Yu2012analysis_kernel}. However, different from these approaches, we train the mode proportion vector $\pi$ in an end-to-end fashion using neural networks and integrate the instance weighting in a Wasserstein optimization. Of more relevance to our work is the method proposed in \cite{Chen2018reweighted}, where the instance weighting is trained end-to-end in a neural network. However, in \cite{Chen2018reweighted}, instance weights are maximized with respect to the Wasserstein loss, while we show that the mixture proportions need to minimized to normalize mode mismatches. Moreover, our NW measure formulation can handle the case when mode assignments for source embeddings are unknown (as we discuss in Section~\ref{sec:UDA_unsup}). This case cannot be handled by the approach presented in \cite{Chen2018reweighted}.

For unsupervised tasks when mode assignments for source samples are unknown, we cannot use the simplified formulation of \eqref{eq:matching-dist}. In that case, we use a domain adaptation method solving the following optimization:
\begin{align}\label{opt:domain-unsupervised-relaxed}
\min_{f\in \cF}~~ &  L_{unsup}\left(X_{s}\right) + \lambda W_N\left(f(X_s),f(X_t)\right) ,
\end{align}
where $L_{unsup}(X_{s})$ is the loss corresponding to the desired {\it unsupervised} learning task on the source domain data.

\subsection{UDA for supervised tasks}\label{sec:UDA_sup}
\subsubsection{MNIST $\to$ MNIST-M}\label{sec:domain-mnist}

In the first set of experiments\footnote{Code available at \url{https://github.com/yogeshbalaji/Normalized-Wasserstein}}, we consider adaptation between MNIST$\to$ MNIST-M datasets. We consider three settings with imbalanced class proportions in source and target datasets: $3$ modes, $5$ modes, and $10$ modes. More details can be found in Table.~\ref{tab:MNIST_settings} of Appendix.

We use the same architecture as ~\cite{ganin2015DANN} for feature network and discriminator. We compare our method with the following approaches: (1) Source-only which is a baseline model trained only on source domain with no domain adaptation performed, (2) DANN~\cite{ganin2015DANN}, a method where adversarial distance between source and target distibutions is minimized, and (3) Wasserstein~\cite{shen2017WDDA} where Wasserstein distance between source and target distributions is minimized. Table~\ref{tab:digits} summarizes our results of this experiment. We observe that performing domain adaptation using adversarial distance and Wasserstein distance leads to decrease in performance compared to the baseline model. This is an outcome of not accounting for mode imbalances, thus resulting in negative transfer, i.e., samples belonging to incorrect classes are coupled and getting pushed to be close in the embedding space. Our proposed NW measure, however, accounts for mode imbalances and leads to a significant boost in performance in all three settings.

\begin{table}[!htb]
\caption{Mean classification accuracies (in $\%$) averaged over $5$ runs on imbalanced MNIST$\to$MNIST-M adaptation}
\label{tab:digits}
\centering
\begin{tabular}{|c|c|c|c|}
\hline
Method & $3$ modes & $5$ modes & $10$ modes \\   
\hline
\hline
Source only &  66.63    &   67.44     &  63.17   \\
DANN       & 62.34 & 57.56 & 59.31 \\
Wasserstein        & 61.75 & 60.56 & 58.22 \\
{\bf NW}        & \textbf{75.06} & \textbf{76.16} & \textbf{68.57} \\
\hline
\end{tabular}
\end{table}

\subsubsection{VISDA}
In the experiment of Section \ref{sec:domain-mnist} on digits dataset, models have been trained from scratch. However, a common practice used in domain adaptation is to transfer knowledge from a pretrained network (eg. models trained on ImageNet) and fine-tune on the desired task. To evaluate the performance of our approach in such settings, we consider adaptation on the VISDA dataset~\cite{visda_challenge}; a recently proposed benchmark for adapting from synthetic to real images. 


We consider a subset of the entire VISDA dataset containing the following three classes: \emph{aeroplane}, \emph{horse} and \emph{truck}. The source domain contains $(0.55, 0.33, 0.12)$ fraction of samples per class, while that of the target domain is $(0.12, 0.33, 0.55)$. We use a Resnet-18 model pre-trained on ImageNet as our feature network. As shown in Table~\ref{tab:visda}, our approach significantly improves the domain adaptation performance over the baseline and other compared methods.

\begin{table}[!htb]
\caption{Mean classification accuracies (in $\%$) averaged over $5$ runs on synthetic to real adaptation on VISDA dataset (3 classes) }
\label{tab:visda}
\centering
\begin{tabular}{|c|c|}
\hline
Method & Accuracy (in $\%$) \\   
\hline
\hline
Source only & 53.19  \\
DANN       &  68.06 \\
Wasserstein  &  64.84 \\
{\bf NW}   & \textbf{73.23} \\
\hline
\end{tabular}
\end{table}

\subsubsection{Mode balanced datasets}
The previous two experiments demonstrated the effectiveness of our method when datasets are imbalanced. In this section, we study the case where source and target domains have mode-balanced datasets -- the standard setting considered in the most domain adaptation methods. We perform experiment on MNIST$\to$MNIST-M adaptation using the entire dataset. Table~\ref{tab:DA_balanced_MNIST} reports the results obtained. We observe that our approach performs on-par with  the standard wasserstein distance minimization. 

\begin{table}[!htb]
\caption{Domain adaptation on mode-balanced datasets: MNIST$\to$MNIST-M. Average classification accuracies averaged over $5$ runs are reported}
\label{tab:DA_balanced_MNIST}
\centering
\begin{tabular}{|c|c|}
\hline
Method & Accuracy (in $\%$) \\   
\hline
\hline
Source only &    60.22    \\
DANN &      85.24  \\
Wasserstein &    83.47    \\
\textbf{NW}        & 84.16  \\ 
\hline
\end{tabular}
\end{table}

\subsection{UDA for unsupervised tasks}\label{sec:UDA_unsup}
\label{sec:DA_unsup}
\begin{figure*}
\centering
\includegraphics[width=\textwidth]{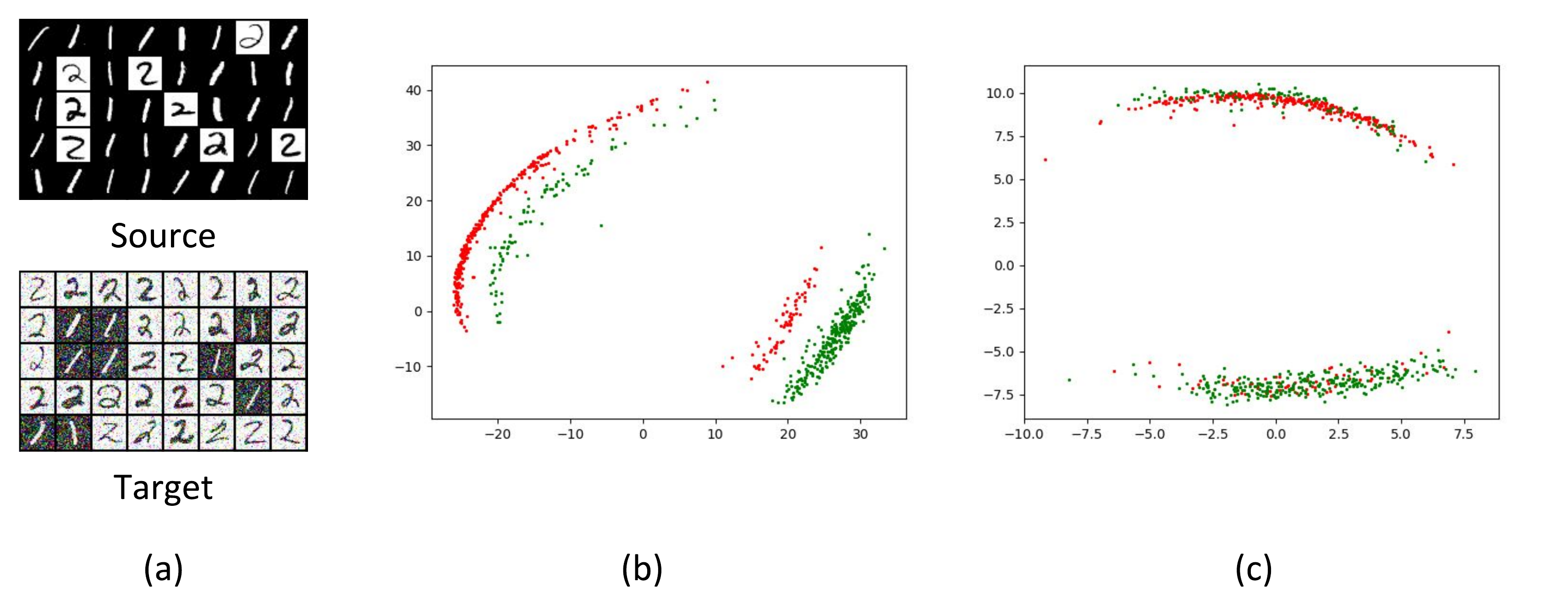}
\caption{Domain adaptation for image denoising. (a) Samples from source and target domains. (b) Source and target embeddings learnt by the baseline model. (c) Source and target embeddings learnt by minimizing the proposed NW measure. In (b) and (c), red and green points indicate source and target samples, respectively.}
\label{fig:DA_unsup} 
\end{figure*}

For unsupervised tasks on mixture datasets, we use the formulation of Eq~\eqref{opt:domain-unsupervised-relaxed} to perform domain adaptation. To empirically validate this formulation, we consider the image denoising problem. The source domain consists of digits $\{1, 2\}$ from MNIST dataset as shown in Fig~\ref{fig:DA_unsup}(a). Note that the color of digit $2$ is inverted. The target domain is a noisy version of the source, i.e. source images are perturbed with random $i.i.d$ Gaussian noise $\mathcal{N}(0.4, 0.7)$ to obtain target images. Our dataset contains $5,000$ samples of digit $1$ and $1,000$ samples of digit $2$ in the source domain, and $1,000$ samples of noisy digit $1$ and $5,000$ samples of noisy digit $2$ in the target. The task is to perform image denoising by dimensionaly reduction, i.e., given a target domain image, we need to reconstruct the corresponding clean image that looks like the source. We assume that no (source, target) correspondence is available in the dataset. 

To perform denoising when the (source, target) correspondence is unavailable, a natural choice would be to minimize the reconstruction loss in source while minimizing the distance between source and target embedding distributions. We use the NW measure as our choice of distance measure. This results in the following optimization:
\begin{align*}
\min_{f, g}~~ & \EE_{\bx \sim X_s} \| g(f(\bx)) - \bx\|_2^2 + \lambda W_N\left(f(X_s),f(X_t)\right) 
\end{align*}
where $f(.)$ is the encoder and $g(.)$ is the decoder. 

As our baseline, we consider a model trained only on source using a quadratic reconstruction loss. Fig~\ref{fig:DA_unsup}(b) shows source and target embeddings produced by this baseline. In this case, the source and the target embeddings are distant from each other. However, as shown in Fig~\ref{fig:DA_unsup}(c), using the NW formulation, the distributions of source and target embeddings match closely (with estimated mode proportions) . We measure the $L_2$ reconstruction loss of the target domain, $err_{recons, tgt} =  \EE_{\bx \sim X_t} \| g(f(\bx)) - \bx\|^2_2$, as a quantitative evaluation measure. This value for different approaches is shown in Table~\ref{tab:err_recons}. We observe that our method outperforms the compared approaches.

\begin{table}[!htb]
\caption{$err_{recons, tgt}$ for an image denoising task  }
\label{tab:err_recons}
\centering
\begin{tabular}{|c|c|}
\hline
Method & $err_{recons, tgt}$ \\   
\hline
\hline
Source only &    0.31    \\
Wasserstein &   0.52  \\
\textbf{NW} & \textbf{0.18}  \\
\hline
Training on target (Oracle) & 0.08 \\
\hline
\end{tabular}
\end{table}

\section{Normalized Wasserstein GAN}\label{sec:gans}
Learning a probability model from data is a fundamental problem in statistics and machine learning. Building on the success of deep learning, a recent approach to this problem is using Generative Adversarial Networks (GANs) \cite{goodfellow2014generative}. GANs view this problem as a {\em game}  between a {\it generator} whose goal is to generate fake samples that are close to the real data training samples, and a {\it discriminator} whose goal is to distinguish between the real and fake samples. 


Most GAN frameworks can be viewed as methods that {\it minimize} a distance between the observed probability distribution, $\PP_X$, and the generative probability distribution, $\PP_{Y}$, where $Y=\bG(Z)$. $\bG$ is referred to as the generator function. In several GAN formulations, the {\it distance} between $\PP_X$ and $\PP_Y$ is formulated as another optimization which characterizes the discriminator. Several GAN architectures have been proposed in the last couple of years. A summarized list includes GANs based on {\bfseries optimal transport} measures (e.g. Wasserstein GAN+Weight Clipping \cite{arjovsky2017wasserstein}, WGAN+Gradient Penalty \cite{gulrajani2017improved}), GANs based on {\bfseries divergence} measures (e.g. the original GAN's formulation \cite{goodfellow2014generative}, DCGAN \cite{radford2015unsupervised}, $f$-GAN \cite{nowozin2016f}), GANs based on {\bfseries moment-matching} (e.g. MMD-GAN \cite{dziugaite2015training,li2015generative}), and other formulations (e.g.  Least-Squares GAN \cite{mao2016multi}, BigGAN \cite{brock2018BIGGAN}, etc.) 
\begin{figure*}
\centering
\includegraphics[width=\textwidth]{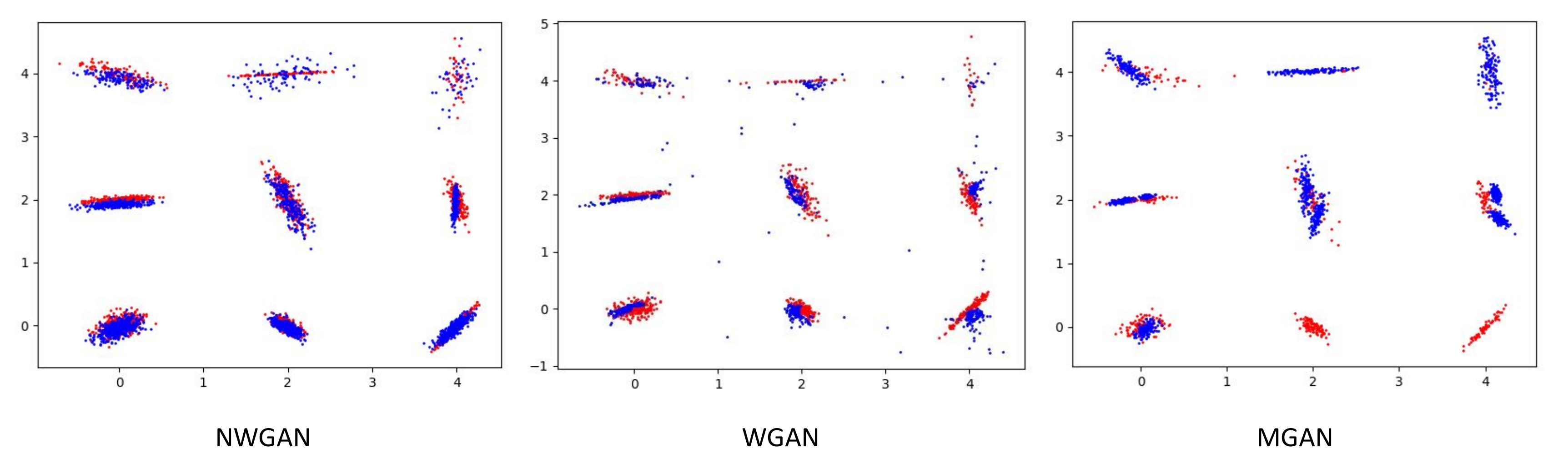}
\caption{Mixture of Gaussian experiments. In all figures, red points indicate samples from the real data distribution while blue points indicate samples from the generated distribution. NWGAN is able to capture rare modes in the data and produces a significantly better generative model than other methods. }
\label{fig:MOG} 
\end{figure*}

If the observed distribution $\PP_X$ is a mixture one, the proposed normalized Wasserstein measure \eqref{eq:N-W} can be used to compute a generative model. Instead of estimating a single generator $\bG$ as done in standard GANs, we estimate a mixture distribution $\PP_{\bG,\pi}$ using the proposed NW measure. We refer to this GAN as the {\it Normalized Wasserstein GAN} (or NWGAN) formulated as the following optimization:
\begin{align}\label{eq:RWGAN}
\min_{\bG,\pi} ~~ &W_N(\PP_X,\PP_{\bG,\pi}).
\end{align}
In this case, the NW distance simplifies as
\begin{align}
&\min_{\bG,\pi} W_N(\PP_X,\PP_{\bG,\pi})\nonumber \\
&= \min_{\bG,\pi} \min_{\bG',\pi^{(1)},\pi^{(2)}} W(\PP_{X},\PP_{\bG',\pi^{(1)}}) +W(\PP_{\bG,\pi}, \PP_{\bG',\pi^{(2)}}) \nonumber \\
 &= \min_{\bG,\pi} W(\PP_{X},\PP_{\bG,\pi}). \label{eq:NWGAN}
\end{align}
There are couple of differences between the proposed NWGAN and the existing GAN architecures. The generator in the proposed NWGAN is a mixture of $k$ models, each producing $\pi_i$ fraction of generated samples. We select $k$ a priori based on the application domain while $\pi$ is computed within the NW distance optimization. Modeling the generator as a mixture of $k$ neural networks has also been investigated in some recent works~\cite{hoang2018mgan, ghosh2017multi}. However, these methods assume that the mixture proportions $\pi$ are known beforehand, and are held fixed during the training. In contrast, our approach is more general as the mixture proportions are also optimized. Estimating mode proportions have several important advantages: (1) we can estimate rare modes, (2) an imbalanced dataset can be re-normalized, (3) by allowing each $\bG_i$ to focus only on one part of the distribution, the quality of the generative model can be improved while the complexity of the generator can be reduced. In the following, we highlight these properties of NWGAN on different datasets.

%
%
%


\subsection{Mixture of Gaussians}
\label{sec:NWGAN_MOG}

First, we present the results of training the NWGAN on a two dimensional mixture of Gaussians. The input data is a mixure of $9$ Gaussians, each centered at a vertex of a $3\times 3$ grid as shown in Figure~\ref{fig:MOG}. The mean and the covariance matrix for each mode are randomly chosen. The mode proportion for mode $i$ is chosen as $\pi_i = \frac{i}{45}$ for $1\leq i\leq 9$. 

Generations produced by NWGAN using $k=9$ affine generator models on this dataset is shown in Figure~\ref{fig:MOG}. We also compare our method with WGAN~\cite{arjovsky2017wasserstein} and MGAN~\cite{hoang2018mgan}. Since MGAN does not optimize over $\pi$, we assume uniform mode proportions ($\pi_i=1/9$ for all $i$). To train WGAN, a non-linear generator function is used since a single affine function cannot model a mixture of Gaussian distribution. 

To evaluate the generative models, we report the following quantitative scores: (1) the average mean error which is the mean-squared error (MSE) between the mean vectors of real and generated samples per mode averaged over all modes, (2) the average covariance error which is the MSE between the covariance matrices of real and generated samples per mode averaged over all modes, and (3) the $\pi$ estimation error which is the normalized MSE between the $\pi$ vector of real and generated samples. Note that computing these metrics require mode assignments for generated samples. This is done based on the closeness of generative samples to the ground-truth means. 

We report these error terms for different GANs in Table~\ref{tab:MOG_quant}. We observe that the proposed NWGAN achieves best scores compared to the other two approaches. Also, from Figure~\ref{fig:MOG}, we observe that the generative model trained by MGAN misses some of the rare modes in the data. This is because of the error induced by assuming fixed mixture proportions when the ground-truth $\pi$ is non-uniform. Since the proposed NWGAN estimates $\pi$ in the optimization, even rare modes in the data are not missed. This shows the importance of estimating mixture proportions specially when the input dataset has imbalanced modes.
\begin{table}[!htb]
\caption{Quantitative Evaluation on Mixture of Gaussians}
\label{tab:MOG_quant}
\centering
\begin{tabular}{|c|c|c|c|}
\hline
Method & Avg. $\mu$ error & Avg. $\Sigma$ error & $\pi$ error \\   
\hline
\hline
WGAN       & 0.007 & 0.0003 & 0.0036 \\
MGAN        & 0.007 & 0.0002 & 0.7157 \\
{\bfseries NWGAN}        & \textbf{0.002} & \textbf{0.0001} & \textbf{0.0001} \\
\hline
\end{tabular}
\end{table}
\subsection{A Mixture of CIFAR-10 and CelebA}
One application of learning mixture generative models is to disentangle the data distribution into multiple components where each component represents one mode of the input distribution. Such disentanglement is useful in many tasks such as clustering (Section~\ref{sec:adv_clustering}). To test the effectiveness of NWGAN in performing such disentanglement, we consider a mixture of $50,000$ images from CIFAR-10 and $100,000$ images from CelebA~\cite{liu2015faceattributes} datasets as our input distribution. All images are reshaped to be $32 \times 32$. 

\begin{figure*}
\centering
\includegraphics[width=\textwidth]{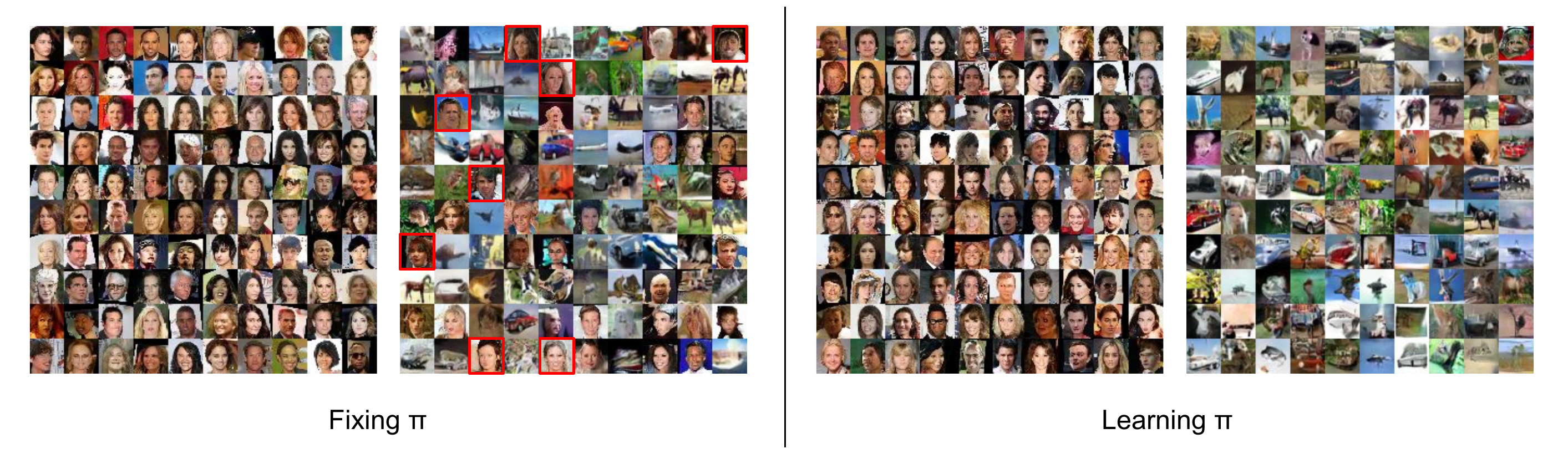}
\caption{Sample generations of NWGAN with $k=2$ on a mixture of CIFAR-10 and CelebA datasets for fixed and optimized $\pi$'s. When $\pi$ is fixed, one of the generators produces a mix of CIFAR and CelebA generative images (boxes in red highlight some of the CelebA generations in the model producing CIFAR+CelebA). However, when $\pi$ is optimized, the model produces disentangled representations.}
\label{fig:CIFAR_celeba} 
\end{figure*}

To highlight the importance of optimizing mixture proportion to produce disentangled generative models, we compare the performance of NWGAN with a variation of NWGAN where the mode proportion $\pi$ is held fixed as $\pi_i=\frac{1}{k}$ (the uniform distribution). Sample generations produced by both models are shown in Figure~\ref{fig:CIFAR_celeba}. When $\pi$ is held fixed, the model does not produce disentangled representations (in the second mode, we observe a mix of CIFAR and CelebA generative images.) However, when we optimize $\pi$, each generator produces distinct modes. 

\section{Adversarial Clustering} \label{sec:adv_clustering}
In this section, we use the proposed NW measure to formulate an adversarial clustering approach. More specifically, let the input data distribution have $k$ underlying modes (each representing a cluster), which we intend to recover. The use of deep generative models for performing clustering has been explored in \cite{yu2018mixtureganclustering} (using GANs) and \cite{locatello2018clustering_generative}(using VAEs). Different from these, our approach makes use of the proposed NWGAN for clustering, and thus explicitly handles data with imbalanced modes. 

Let $\PP_X$ be observed empirical distribution. Let $\bG^*$ and $\pi^*$ be optimal solutions of NWGAN optimization \eqref{eq:NWGAN}. For a given point $\bx_i \sim \PP_X$, the clustering assignment is computed using the closest distance to a mode i.e.,
\begin{align}
C(\bx_i)=\arg\min_{1\leq j\leq k} ~ \min_{Z}\left[\|\bx_i-\bG_j(Z)\|^2\right].
\end{align}
To perform an effective clustering, we require each mode $\bG_j$ to capture one mode of the data distribution. Without enforcing any regularization and using rich generator functions, one model can capture multiple modes of the data distribution. To prevent this, we introduce a {\it regularization} term that maximizes the weighted average Wasserstein distances between different generated modes. That is,
\begin{align}\label{eq:reg}
R = \sum_{(i, j)| i>j} \pi_i \pi_j W\left(\bG_{i}(Z), \bG_{j}(Z)\right).
\end{align}
This term encourages {\it diversity} among generative modes. With this regularization term, the optimization objective of a {\it regularized} NWGAN becomes
\begin{align*}
\min_{\bG,\pi} W(\PP_{X},\PP_{\bG,\pi}) - \lambda_{reg} R
\end{align*}
where $\lambda_{reg}$ is the regularization parameter. 

We test the proposed adversarial clustering method on an imbalanced MNIST dataset with $3$ digits containing $3,000$ samples of digit $2$, $1,500$ samples of digit $4$ and $6,000$ samples of digit $6$. We compare our approach with \emph{k-means} clustering and \emph{Gaussian Mixture Model (GMM)} in Table~\ref{tab:clustering}. Cluster purity, NMI and ARI scores are used as quantitative metrics (refer to Appendix~\ref{app:clustering_metrics} for more details). We observe that our clustering technique is able to achieve good performance over the compared approaches.

\begin{table}[!htb]
\caption{Clustering results on Imbalanced MNIST dataset }
\label{tab:clustering}
\centering
\begin{tabular}{|c|c|c|c|}
\hline
Method & Cluster Purity & NMI & ARI \\   
\hline
\hline
k-means   & 0.82 & 0.49 & 0.43 \\
GMM        & 0.75 & 0.28 & 0.33 \\
{\bf NW} & \textbf{0.98} & \textbf{0.94} & \textbf{0.97}  \\
\hline
\end{tabular}
\end{table}

\section{Choosing the number of modes}\label{sec:choosing_num_modes}
\begin{figure}
\centering
\includegraphics[width=0.4\textwidth]{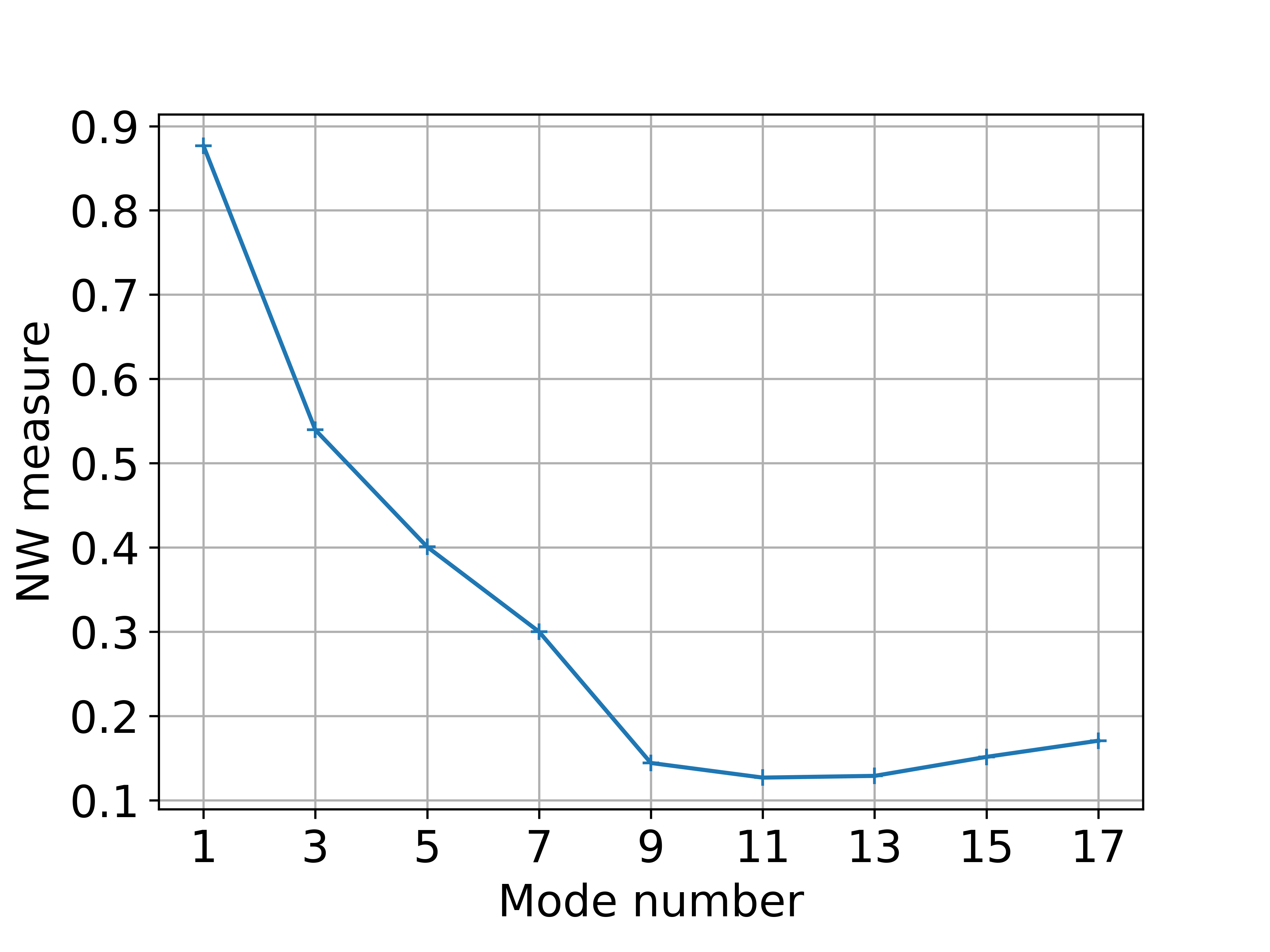}
\caption{Choosing $k$: Plot of NW measure vs number of modes}
\label{fig:sensitivity} 
\end{figure}

As discused in Section~\ref{sec:theory}, choosing the number of modes ($k$) is crucial for computing NW measure. While this information is available for tasks such as domain adaptation, it is unknown for others like generative modeling. In this section, we experimentally validate our theoretically justified algorithm for estimating $k$.
Consider the mixture of Gaussian dataset with $k=9$ modes presented in Section~\ref{sec:NWGAN_MOG}.  On this dataset, the NWGAN model (with same architecture as that used in Section~\ref{sec:NWGAN_MOG}) was trained with varying number of modes $k$. For each setting, the NW measure between the generated and real data distribution is computed and plotted in Fig~\ref{fig:sensitivity}. We observe that $k=9$ satisfies the condition discussed in Theorem~\ref{thm:main}: optimal $k^*$ is the smallest $k$ for which $NW(k)$ is small, $NW(k-1) - NW(k)$ is large, and $NW(k)$ saturates after $k^*$.

\section{Conclusion}
In this paper, we showed that Wasserstein distance, due to its marginal constraints, can lead to undesired results when when applied on imbalanced mixture distributions. To resolve this issue, we proposed a new distance measure called the Normalized Wasserstein. The key idea is to optimize mixture proportions in the distance computation, effectively normalizing mixture imbalance. We demonstrated the usefulness of NW measure in three machine learning tasks: GANs, domain adaptation and adversarial clustering. Strong empirical results on all three problems highlight the effectiveness of the proposed distance measure. 

\section{Acknowledgements}
Balaji and Chellappa were supported by MURI program from the Army Research Office (ARO) under the grant W911NF17-1-0304. Feizi was supported by the US National Science Foundation (NSF) under the grant CDS\&E:1854532, and Capital One Services LLC.

\bibliographystyle{ieee_fullname}
\bibliography{refs}

\setcounter{theorem}{0}
\newpage
\section*{Appendices}
\begin{appendix}

\section{Proof of Theorem 1}\label{app:proof}

For NW measure to normalize mode proportions appropriately, we need a good estimate of the number of mode proportions. Theorem 1 provides conditions under which the mode proportions can provable be estimated. 

Let $\mathbb{P}_X$ and $\mathbb{P}_Y$ be two mixture distributions whose NW measure we wish to compute. Let $\mathbb{P}_X$ and $\mathbb{P}_Y$ have $n_1$ and $n_2$ modes respectively, with $r$ modes overlapping. Let $k^* = n_1 + n_2 - r$. We make the following assumptions
\begin{itemize}
    \item (A1) If mode $i$ in distribution $X$ and mode $j$ in distribution $Y$ belong to the same mixture component, then their Wasserstein distance is $\leq \epsilon$ i.e., if $X_i$ and $Y_j$ correspond to the same component, $W(\PP_{X_i}, \PP_{Y_j}) < \epsilon$.
    \item (A2) The minimum Wasserstein distance between any two modes of one mixture distribution is at least $\delta$ i.e., $W(\PP_{X_i}, \PP_{X_j}) > \delta$ and $W(\PP_{Y_i}, \PP_{Y_j}) > \delta$ $\forall i \neq j$. Also, non-overlapping modes between $X$ and $Y$ are separated by $\delta$ i.e., for non-overlapping modes $X_i$ and $Y_j$,  $W(\PP_{X_i}, \PP_{Y_j}) > \delta$.  This ensures that modes are well-separated.
    \item (A3) We assume that each mode $X_i$ and $Y_i$ have density at least $\eta$ i.e., $\PP_{X_i} \geq \eta \quad \forall i$, $\PP_{Y_i} \geq \eta \quad \forall i$. This ensures that every mode proportion is at least $\eta$.
    \item (A4) Each generator $\bG_i$ is powerful enough to capture exactly one mode of distribution $\PP_X$ or $\PP_Y$.
\end{itemize}
\begin{lemma}\label{lem:mon}
$NW(k)$ is a monotonically decreasing function with respect to $k$.
\end{lemma}
This is because in $NW(k+1)$, we add one additional mode compared to $NW(k)$. If we have $\pi^{(1)}$, $\pi^{(2)}$ for this new mode to be $0$ and give the same assignements as $NW(k)$ to the rest of the modes, $NW(k+1)=NW(k)$. Since computing $NW(k)$ contains a minimization over mode assignments, the $NW(k+1)\leq NW(k) \forall k$. Hence, it is monotonically decreasing.   

\begin{lemma}\label{lem:NW_kstar}
$NW(k^*) \leq \eps$
\end{lemma}

This is because at $k=k^*$, we can make the following mode assignments. 
\begin{itemize}
    \item Assign $n_1 + n_2 - r$ modes of NW to each of $n_1 + n_2 - r$ non-overlapping modes in $\mathbb{P}_X$ and $\mathbb{P}_Y$ with the same mixture .
    \item Assign the remaining $r$ modes of NW to the overlapping modes of either $\mathbb{P}_X$ or $\mathbb{P}_Y$. WLOG, let us assume we assign them to $r$ overlapping modes of $\mathbb{P}_X$.
    \item Choose $\pi^{(1)}$ to be same as $\pi$ for $\mathbb{P}_X $, with $0$ to non-overlapping components of $\mathbb{P}_Y$
    \item Choose $\pi^{(2)}$ to be same as $\pi$ for $\mathbb{P}_Y $, with $0$ to non-overlapping components of $\mathbb{P}_X$
\end{itemize}

Let us denote $NOv(X)$ to be non-overlapping modes of $X$, $Ov(X)$ to be overlapping modes of $X$, $NOv(Y)$ to be non-overlapping modes of $Y$, and $Ov(Y)$ to be overlapping modes of $Y$. Then, under the mode assignments given above, $NW(k^*)$ can be evaluated as,
\begin{align*}
&W_N(\PP_{X},\PP_{Y})\\
&:=\min_{\bG,\pi^{(1)},\pi^{(2)}} ~W(\PP_{X},\PP_{\bG,\pi^{(1)}})+W(\PP_{Y},\PP_{\bG,\pi^{(2)}}).\nonumber \\
&= \sum_{i \in NOv(X)} \pi_{i}^{X} W(\PP_{X_i},\PP_{X_i}) + \sum_{i \in Ov(X)}\pi_{i}^{X} W(\PP_{X_i},\PP_{X_i})+ \\
& \sum_{i \in NOv(Y)} \pi_{i}^{Y} W(\PP_{Y_i},\PP_{Y_i}) + \sum_{i \in Ov(Y)} \pi_{i}^{Y} W(\PP_{Y_i},\PP_{X_i}) \\
&= 0 + 0 + 0 + \sum_{i \in Ov(Y)} \pi_{i}^{Y} W(\PP_{Y_i},\PP_{X_i}) \\
&\leq \eps
\end{align*}
The last step follows from (A1) i.e., overlapping modes are separated by a Wasserstein distance of $\eps$.

\begin{lemma}\label{lem:NW_kstar_m1}
$NW(k^*-1) \geq \frac{\delta}{2} \eta$
\end{lemma}
By assumption (A2), we know that any two modes have separation of at least $\delta$. In the distribution $\mathbb{P}_X + \mathbb{P}_Y$, there are $n_1+n_2-r$ unique cluster centers, each pair of clusters at a Wasserstein distance $\delta$ distance apart. In $NW(k^* - 1)$, generators have $n_1+n_2-r-1$ modes, which is $1$ less than the number of modes in $\mathbb{P}_X + \mathbb{P}_Y$. 
Now, let us assume that $NW(k^*-1) < \frac{\delta}{2} \eta$. Then, 
\begin{align*}
    W(\PP_X, \PP_{\bG, \pi^{(1)}}) + W(\PP_Y, \PP_{\bG, \pi^{(2)}}) < \frac{\delta}{2}\eta
\end{align*}
Since each mode of $\PP_X$ and $\PP_Y$ has density at least $\eta$ (by (A3)), the above condition can be satisfied only if
\begin{align}
    \forall i \in [n_1], \exists j \in [k^*-1] \text{ s.t. } W(\PP_{X_i}, \PP_{\bG_j}) < \frac{\delta}{2} \label{eq:cond1} \\
    \forall i \in [n_2], \exists j \in [k^*-1] \text{ s.t. } W(\PP_{Y_i}, \PP_{\bG_j}) < \frac{\delta}{2} \label{eq:cond2}
\end{align}
Accounting for $r$ mode overlap between $X$ and $Y$, there will be $n_1+n_2-r$ unique constraints in Eq.~\eqref{eq:cond1} and Eq.~\eqref{eq:cond2}. Since, $\bG$ has only $k^*-1$ modes, by Pigeonhole principle, there should be at least one pair $(i, j)$ that is matched to the same $\bG_j$. WLOG, let us consider both $i$ and $j$ to belong to $\PP_X$, although each can either belong to $\PP_X$ or $\PP_Y$. Then,
\begin{align*}
    W(\PP_{X_i}, \bG_k) &< \frac{\delta}{2} \\
    W(\PP_{X_j}, \bG_k) &< \frac{\delta}{2} 
\end{align*}
Then, by triangle inequality, $W(\PP_{X_i}, \PP_{X_j}) < \delta$. This contradicts assumption (A2). Hence $NW(k^*-1) \geq \frac{\delta}{2} \eta$

\begin{theorem}\label{app_thm:main}
Let $\mathbb{P}_X$ and $\mathbb{P}_Y$ be two mixture distributions satisfying (A1)-(A4) with $n_{1}$ and $n_{2}$ mixture components, respectively, where $r$ of them are overlapping. Let $k^*=n_1 + n_2 - r$. Then, $k^*$ is smallest $k$ for which $NW(k)$ is small ($O(\epsilon)$) and $NW(k) - NW(k-1)$ is relatively large (in the $O(\delta \eta)$ )
\end{theorem}

\paragraph{Proof:} 
From Lemma~\ref{lem:NW_kstar} and Lemma~\ref{lem:mon}, we know that $NW(k) \leq \eps \quad \forall k\geq k^*$. Similarly, from Lemma~\ref{lem:NW_kstar_m1} and Lemma~\ref{lem:mon}, we $NW(k) \geq \frac{\delta}{2} \eta \quad \forall k < k^*$. Hence, $k^*$ is the smallest $k$ for which $NW(k)$ is small ($O(\epsilon)$) and $NW(k) - NW(k-1)$ is relatively large (in the $O(\delta \eta)$ ). Hence, proved.

\section{Properties of Normalized Wasserstein measure}\label{app:NW_properties}
The defined NW measure is not a distance because it does not satisfy the properties of a distance measure.

\begin{itemize}
\item In general, $ W_N(\PP_X, \PP_Y) \neq 0 $. However, if $\PP_X \in \PP_{\bG, \pi}$,  $ W_N(\PP_X, \PP_X) = 0 $. Moreover, if $\exists \bG, \pi \text{ s.t. } W_N(\PP_{\bG, \pi}, \PP_X) < \eps$ (i.e., $\PP_{\bG, \pi}$ approximates $\PP_X$ within $\eps$ factor), then $ W_N(\PP_X, \PP_X) \leq 2\eps $. This follows from the definition of NW measure. So, when the class of generators are powerful enough, this property is satisfied within $2\eps$ approximation

\item Normalized Wasserstein measure is symmetric. $W_N(\PP_X, \PP_Y) = W_N(\PP_Y, \PP_X)$

\item Normalized Wasserstein measure does not satisfy triangle inequality.

\end{itemize}

\section{Optimizing Normalized Wasserstein using duality}\label{app:duality}

NW measure between two distributions $\PP_X$ and $\PP_Y$ is defined as 
\begin{align*}
\min_{\bG,\pi^{(1)},\pi^{(2)}} ~ &W(\PP_{X},\PP_{\bG,\pi^{(1)}})+W(\PP_{Y},\PP_{\bG,\pi^{(2)}})
\end{align*}
Similar to ~\cite{arjovsky2017wasserstein}, using the dual of Wasserstein distance, we can write the above optimization as 
\begin{align}\label{eq:NW_dual}
\min_{\bG,\pi^{(1)},\pi^{(2)}} ~ \Big[ & \max_{D_1 \in 1-Lip} \EE[D_1(X)] - \EE[\sum_i \pi_i^{(1)} D_1(G_i(Z))] + \nonumber \\
        & \max_{D_2 \in 1-Lip} \EE[D_2(Y)] - \EE[\sum_i \pi_i^{(2)} D_2(G_i(Z))] \Big]
\end{align}

Here, $D_1$ and $D_2$ are 1-Lipschitz functions, and $\pi^{(1)}$ and $\pi^{(2)}$ are $k-$dimesional vectors lying in a simplex i.e., 
\begin{align*}
\sum_i \pi^{(1)}_i = 1, \quad \sum_i \pi^{(2)}_i = 1
\end{align*}
To enforce these constraints, we use the softmax function as follows. 
\begin{align*}
\pi^{(1)} = \text{softmax}(\tilde{\pi}^{(1)}), \quad \pi^{(2)} = \text{softmax}(\tilde{\pi}^{(2)})
\end{align*}
The new variables $\tilde{\pi}^{(1)}$ and $\tilde{\pi}^{(2)}$ become optimization variables. The softmax function ensures that the mixture probabilities $\pi^{(1)}$ and $\pi^{(2)}$ lie in a simplex.

The above equations are optimized using alternating gradient descent given by the following algorithm.
\begin{algorithm}[!htb]
\caption{Optimizatizing Normalized Wasserstein}\label{alg:NW}
\begin{algorithmic}[1]
\State Training iterations = $N_{iter}$
\State Critic iterations = $N_{critic}$
\For{$t = 1:N_{iter}$}
		\State Sample minibatch $\bx \sim \PP_X$, $\by \sim \PP_Y$ 
		\State Sample minibatch $\bz \sim \cN(0, 1)$
	    \State Compute Normalized Wasserstein as 
	            \begin{align*}
	                NW  &= \EE[D_1(\bx)] - \EE[\sum_i \pi_i^{(1)} D_1(G_i(\bz))] + \\
	                    & \EE[D_2(\by)] - \EE[\sum_i \pi_i^{(2)} D_2(G_i(\bz))]
	            \end{align*}
		\For{$k = 1:N_{critic}$}
            \State Maximize $NW$ w.r.t $D_1$ and $D_2$ 
            \State Minimize $NW$ w.r.t $\tilde{\pi}^{(1)}$ and $\tilde{\pi}^{(2)}$ 
        \EndFor
        \State Minimize $NW$ w.r.t $G$ 
\EndFor
\end{algorithmic}
\end{algorithm}

\section{Comparative analysis of mixture distributions}\label{sec:hypothesis-testing} 

In this section, we propose a test using a combination of Wasserstein distance and NW measure to identify if two mixture distributions differ in mode components or mode proportions. Such a test can provide better understanding while comparing mixture distributions. Suppose $\PP_{X}$ and $\PP_{Y}$ are two mixture distributions with the same mixture components but different mode proportions. I.e., $\PP_{X}$ and $\PP_{Y}$ both belong to $\cP_{\bG,k}$. In this case, depending on the difference between $\pi^{(1)}$ and $\pi^{(2)}$, the Wasserstein distance between the two distributions can be arbitrarily large. Thus, using the Wasserstein distance, we can only conclude that the two distributions are different. In some applications, it can be informative to have a test that determines if two distributions differ only in mode proportions. We propose a test based on a combination of Wasserstein and the NW measure for this task. This procedure is shown in Table.~\ref{tab:hyp_test}. We note that computation of $p$-values for the proposed test is beyond the scope of this paper.

\begin{table*}
\caption{Comparative analysis of two mixture distributions  }
\label{tab:hyp_test}
\centering
\begin{tabular}{|c|c|c|}
\hline
\shortstack{Wasserstein \\ distance} & \shortstack{{NW} \\ {measure}} & Conclusion \\   
\hline
\hline
High & High & Distributions differ  in mode components   \\
\hline
High & Low & \shortstack{Distributions have the same components, \\ but differ in mode proportions}     \\
\hline
Low  & Low & Distributions are  the same \\
\hline
\end{tabular}
\end{table*}

We demonstrate this test on 2D Mixture of Gaussians. We perform experiments on two settings, each involving two datasets $\mathcal{D}_1$ and $\mathcal{D}_2$, which are mixtures of $8$ Gaussians:

{\bf Setting 1:}  Both $\mathcal{D}_{1}$ and $\mathcal{D}_{2}$ have same mode components, with the $i^{th}$ mode located at $(r \cos(\frac{2\pi i}{8}), r \sin(\frac{2\pi i}{8}))$. 

{\bf Setting 2:}  $\mathcal{D}_{1}$ and $\mathcal{D}_{2}$ have shifted mode components. The $i^{th}$ mode of $\mathcal{D}_{1}$  is located at $(r \cos(\frac{2\pi i}{8}), r \sin(\frac{2\pi i}{8}))$, while the $i^{th}$ mode of $\mathcal{D}_{2}$ is located at $(r \cos(\frac{2\pi i + \pi}{8}), r \sin(\frac{2\pi i + \pi}{8}))$. 

In both the settings, the mode fraction of $\mathcal{D}_1$ is $\pi_i=\frac{i+2}{52}$, and that of $\mathcal{D}_2$ is $\pi_i=\frac{11-i}{52}$. We use $2,000$ data points from $\mathcal{D}_{1}$ and $\mathcal{D}_{2}$ to compute Wasserstein distance and the NW measure in primal form by solving a linear program. The computed distance values are reported in Table~\ref{tab:hyp_MOG}. In setting 1, we observe that the Wasserstein distance is large while the NW measure is small. Thus, one can conclude that the two distributions differ only in mode proportions. In setting 2, both Wasserstein and NW measures are large. Thus, in this case, distributions differ in mixture components as well.

\begin{table}[!htb]
\caption{Hypothesis test between two MOG - $\mathcal{D}_{1}$ and $\mathcal{D}_{2}$ }
\label{tab:hyp_MOG}
\centering
\begin{tabular}{|c|c|c|}
\hline
Setting & Wasserstein Distance & NW measure \\   
\hline
\hline
Setting 1 & 1.51 & 0.06 \\
Setting 2 & 1.56 & 0.44 \\
\hline
\end{tabular}
\end{table}

\section{Additional results}

\subsection{CIFAR-10}

\begin{figure*}
\centering
\includegraphics[width=\textwidth]{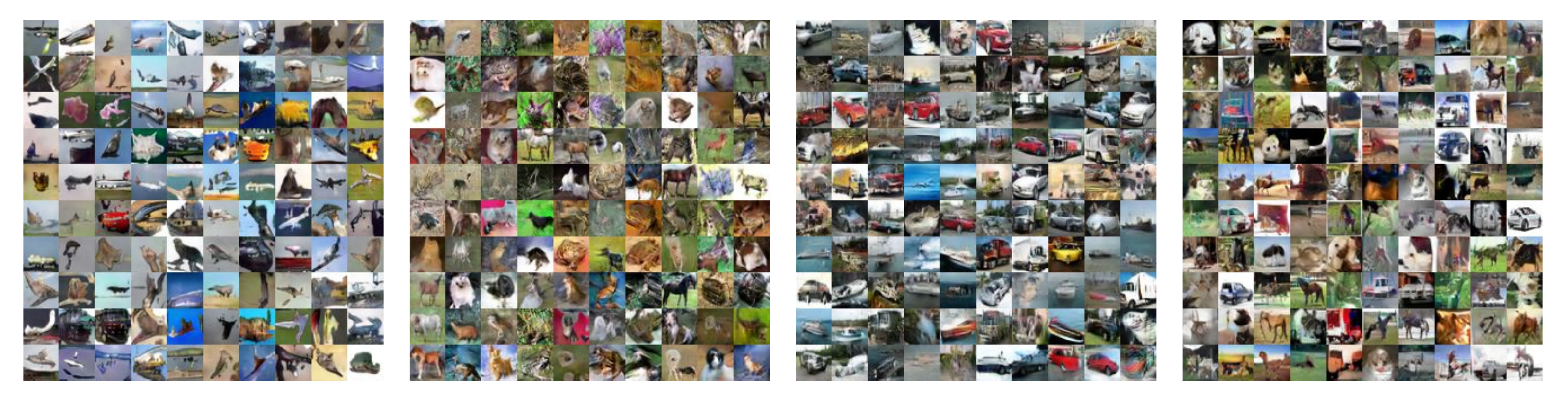}
\caption{Sample generations produced by the proposed NWGAN trained on CIFAR-10 with $k=4$ generator modes.}
\label{fig:CIFAR} 
\end{figure*}

We present the results of training NWGAN on CIFAR-10 dataset. We use WGAN-GP~\cite{gulrajani2017improved} with Resnet-based generator and discriminator models as our baseline method. The proposed NWGAN was trained with $k=4$ modes using the same network architectures as the baseline. Sample generations produced by each mode of the NWGAN is shown in Figure~\ref{fig:CIFAR}. We observe that each generator model captures distinct variations of the entire dataset, thereby approximately disentangling different modes in input images. For quantitative evaluation, we compute inception scores for the baseline and the proposed NWGAN. The inception score for the baseline model is $7.56$, whereas our model achieved an improved score of $7.89$.


\begin{table*}
\caption{MNIST $\to$ MNIST-M settings}
\label{tab:MNIST_settings}
\centering
\begin{tabular}{|c|c|c|c|}
\hline
Config & 3 modes & 5 modes & 10 modes \\   
\hline
\hline
Classes       & $\{ 1, 4, 8\}$ & $\{0, 2, 4, 6, 8\}$ & $\{ 0, 1, \hdots 9 \}$ \\ \hline
\shortstack{Proportion of \\ source samples}        & \shortstack{$\{ 0.63, 0.31,$ \\$ 0.06 \}$} & \shortstack{$\{ 0.33, 0.26, 0.2, $ \\$ 0.13, 0.06 \}$} & \shortstack{$\{ 0.15, 0.15, 0.15, 0.12, 0.12 $ \\$ 0.11, 0.05, 0.05, 0.05, 0.05 \}$} \\ \hline
\shortstack{Proportion of \\ target samples}        & \shortstack{$\{ 0.06, 0.31,$ \\$ 0.63 \}$} & \shortstack{$\{ 0.06, 0.13, 0.2, $ \\$ 0.26, 0.33 \}$} &  \shortstack{$\{ 0.05, 0.05, 0.05, 0.05, 0.11 $ \\$ 0.12, 0.12, 0.15, 0.15, 0.15 \}$} \\
\hline
\end{tabular}
\end{table*}

\subsection{Domain adaptation under uniform mode proportions}
In this section, we present results on domain adaptation on mode-balanced VISDA dataset --  source and target domains contain $3$ classes - \emph{aeroplane}, \emph{horse} and \emph{truck}  with uniform mode proportion. The results of performing adaptation using NW measure in comparison with classical distance measures are reported in Table~ \ref{tab:DA_balanced_VISDA}. We observe that NW measure performs on-par with the compared methods on this dataset. This experiment demonstrates the effectiveness of NW measure on a range of settings -- when the source and target datasets are balanced in mode proportions, NW becomes equivalent to Wasserstein distance and minimizing it is no worse than minimizing the classical distance measures. On the other hand, when mode proportions of source and target domains differ, NW measure renormalizes the mode proportions and effectively performs domain adaptation. This illustrates the usefulness of NW measure in domain adaptation problems.

\begin{table}[!htb]
\caption{Domain adaptation on mode-balanced datasets: VISDA. Average classification accuracies averaged over $5$ runs are reported}
\label{tab:DA_balanced_VISDA}
\centering
\begin{tabular}{|c|c|}
\hline
Method & Classification accuracy (in $\%$) \\   
\hline
\hline
Source only &    63.24   \\
DANN &     84.71   \\
Wasserstein &   90.08     \\
Normalized Wasserstein        & 90.72  \\ 
\hline
\end{tabular}
\end{table}

\subsection{Adversarial clustering: Quantitative metrics}\label{app:clustering_metrics}
\begin{itemize}
\item Cluster purity: Cluster purity measures the extent to which clusters are consistent i.e., if each cluster constains similar points or not. To compute the cluster purity, the cardinality of the majority class is computed for each cluster, and summed over and divided by the total number of samples.

\item ARI - Adjusted Rand Index: The rand index computes the similarity measure between two clusters by considering all pairs of samples, and counting the pairs of samples having the same cluster in the ground-truth and predicted cluster assignments. Adjusted rand index makes sure that ARI score is in the range (0, 1)

\item NMI - Normalized Mutual Information: NMI is the normalized version of the mutual information between the predicted and the ground truth cluster assignments.

\end{itemize}

\subsection{Adversarial clustering of CIFAR+CelebA}
In this section, we show the results of performing adversarial clustering on a mixture of CIFAR-10 and CelebA datasets. The same dataset presented in Section 3.2 of the main paper is used in this experiment (i.e) the dataset contains CIFAR-10 and CelebA samples in $1:2$ mode proportion. NWGAN was trained with $2$ modes - each employing Resnet based generator-discriminator architectures (same architectures and hyper-parameters used in Section 3.2 of main paper). Quantitative evaluation of our approach in comparison with $k-means$ is given in Table~\ref{tab:clustering_CIFAR}. We observe that our approach outperforms $k-means$ clustering. However, the clustering quality is poorer that the one obtained on imbalanced MNIST dataset. This is because the samples generated on MNIST dataset had much better quality than the one produced on CIFAR-10. So, as long as the underlying GAN model produces good generations, our adversarial clustering algorithm performs well.

\begin{table}[!htb]
\caption{Performance of clustering algorithms on CIFAR+CelebA dataset }
\label{tab:clustering_CIFAR}
\centering
\begin{tabular}{|c|c|c|c|}
\hline
Method & Cluster Purity & NMI & ARI \\   
\hline
\hline
k-means   & 0.667 & 0.038 & 0.049  \\
Normalized Wasserstein & 0.870 & 0.505 & 0.547  \\
\hline
\end{tabular}
\end{table}

\section{Architecture and hyper-parameters}

Implementation details including model architectures and hyperparameters are presented in this section:

\subsection{Mixture models for Generative Adversarial Networks (GANs)}

\subsubsection{Mixture of Gaussians}
As discussed in Section 3.1 of the main paper, the input dataset is a mixture of $8$ Gaussians with varying mode proportion. Normalized Wasserstein GAN was trained with linear generator and non-linear discriminator models using the architectures and hyper-parameters as presented in Table~\ref{tab:MOG_relaxed_WGAN}. The architecture used for training Vanilla WGAN is provided in Table~\ref{tab:MOG_WGAN}. The same architecture is used for MGAN, however we do not use the $ReLU$ non-linearities in the Generator function (to make the generator affine so that the model is comparable to ours). For WGAN and MGAN, we use the hyper-parameter details as provided in the respective papers -- \cite{gulrajani2017improved} and \cite{hoang2018mgan}.

\begin{table*}
\caption{Architectures and hyper-parameters: Mixture of Gaussians with Normalized Wasserstein GAN}
\label{tab:MOG_relaxed_WGAN}
\centering
\begin{tabular}{|c|c|}
\hline
Generator & Discriminator\\   
\hline
\hline
Linear($2 \to 64$)  & Linear($2 \to 128$) \\
Linear($64 \to 64$) & LeakyReLU(0.2) \\
Linear($64 \to 64$) & Linear($128 \to 128$) \\
Linear($64 \to 2$)   & LeakyReLU(0.2) \\
 								& Linear($128 \to 2$) \\
\hline
\multicolumn{2}{|c|}{Hyperparameters} \\
\hline
\hline
Discriminator learning rate & $0.00005$ \\
Generator learning rate & $0.00005$ \\
$\pi$ learning rate & $0.01$ \\
Batch size & $1024$ \\
Optimizer & RMSProp \\
Number of critic iters & $10$ \\
Weight clip & $[-0.003, 0.003]$ \\
\hline
\end{tabular}
\end{table*}

\begin{table*}
\caption{Architectures: Mixture of Gaussians with vanilla WGAN model}
\label{tab:MOG_WGAN}
\centering
\begin{tabular}{|c|c|}
\hline
Generator & Discriminator\\   
\hline
\hline
Linear($2 \to 512$) + ReLU &Linear($2 \to 512$) + ReLU\\
Linear($512 \to 512$) +  ReLU & Linear($512 \to 512$) +  ReLU \\
Linear($512 \to 512$) + ReLU& Linear($512 \to 512$) +  ReLU \\
Linear($512 \to 2$)   & Linear($512 \to 2$) \\
\hline
\end{tabular}
\end{table*}

\subsubsection{CIFAR-10 + CelebA}

To train models on CIFAR-10 + CelebA dataset (Section 3.2 of the main paper), we used the Resnet architectures of WGAN-GP~\cite{gulrajani2017improved} with the same hyper-parameter configuration for the generator and the discriminator networks. In Normalized WGAN, the learning rate of mode proportion $\pi$ was $5$ times the learning rate of the discriminator. 

\subsection{Domain adaptation for mixture distributions}

\subsubsection{Digit classification}

For MNIST$\to$MNIST-M experiments (Section 4.1.1 of the main paper), following ~\cite{ganin2015DANN}, a modified Lenet architecture was used for feature network, and a MLP network was used for domain classifier. The architectures and hyper-parameters used in our method are given in Table~\ref{tab:DA_digits}. The same architectures are used for the compared approaches - Source only, DANN and Wasserstein.

\begin{table*}
\caption{Architectures and hyper-parameters: Domain adaptation for MNIST$\to$MNIST-M experiments}
\label{tab:DA_digits}
\centering
\begin{tabular}{|c|c|}
\hline
\multicolumn{2}{|c|}{Feature network} \\   
\hline
\hline
\multicolumn{2}{|c|}{Conv($3\to 32$, $5\times 5$ kernel) + ReLU + MaxPool(2)} \\
\multicolumn{2}{|c|}{Conv($32\to 48$, $5\times 5$ kernel) + ReLU + MaxPool(2)} \\
\hline
Domain discriminator & Classifier \\
\hline
\hline
Linear($768 \to 100$) + ReLU  & Linear($768 \to 100$) + ReLU \\
Linear($100 \to 1$) & Linear($100 \to 100$) + ReLU \\
								 & Linear($100 \to 10$) \\
\hline
\multicolumn{2}{|c|}{Hyperparameters} \\
\hline
\hline
Feature network learning rate &  $0.0002$   \\
Discriminator learning rate & $0.0002$ \\
Classifier learning rate & $0.0002$ \\
$\pi$ learning rate & $0.0005$ \\
Batch size & $128$ \\
Optimizer & Adam \\
Number of critic iters & $10$ \\
Weight clipping value & $[-0.01, 0.01]$\\
$\lambda$ & $1$ \\
\hline
\end{tabular}
\end{table*}

\subsubsection{VISDA}
For the experiments on VISDA dataset with three classes (Section 4.1.2 of the main paper), the architectures and hyper-parameters used in our method are given in Table~\ref{tab:DA_VISDA}. The same architectures are used for the compared approaches: source only, Wasserstein and DANN.

\begin{table*}
\caption{Architectures and hyper-parameters: Domain adaptation on VISDA dataset}
\label{tab:DA_VISDA}
\centering
\begin{tabular}{|c|c|}
\hline
\multicolumn{2}{|c|}{Feature network} \\
\hline
\hline
\multicolumn{2}{|c|}{Resnet-18 model pretrained on ImageNet} \\
\multicolumn{2}{|c|}{till the penultimate layer} \\
\hline
Domain discriminator & Classifier \\
\hline
\hline
Linear($512 \to 512$) + LeakyReLU($0.2$)  & Linear($512 \to 3$) \\
Linear($512 \to 512$) + LeakyReLU($0.2$)  &  \\
Linear($512 \to 512$) + LeakyReLU($0.2$)  &  \\
Linear($512 \to 1$)  &  \\

\hline
\multicolumn{2}{|c|}{Hyperparameters} \\
\hline
\hline
Feature network learning rate &  $0.000001$   \\
Discriminator learning rate & $0.00001$ \\
Classifier learning rate & $0.00001$ \\
$\pi$ learning rate & $0.0001$ \\
Batch size & $128$ \\
Optimizer & Adam \\
Number of critic iters & $10$ \\
Weight clipping value & $[-0.01, 0.01]$\\
$\lambda$ & $1$ \\
\hline
\end{tabular}
\end{table*}

\subsubsection{Domain adaptation for Image denoising}

The architectures and hyper-parameters used in our method for image denoising experiment (Section 4.2 of the main paper) are presented in Table~\ref{tab:DA_denoising}. To perform adaptation using Normalized Wasserstein measure, we need to train the intermediate distributions $\PP_{\bG, \pi^{(1)}}$ and $\PP_{\bG, \pi^{(2)}}$ (as discussed in Section 2, 4.2 of the main paper). We denote the generator and discriminator models corresponding to $\PP_{\bG, \pi^{(1)}}$ and $\PP_{\bG, \pi^{(2)}}$ as Generator (RW) and Discriminator (RW) respectively. In practice, we noticed that the Generator (RW) and Discriminator (RW) models need to be trained for a certain number of iterations first (which we call initial iterations) before performing adaptation. So, for these initial iterations, we set the adaptation parameter $\lambda$ as $0$. Note that the encoder, decoder, generator (RW) and discriminator (RW) models are trained during this phase, but the adaptation is not performed. After these initial iterations, we turn the adaptation term on. The hyperparameters and model architectures are given in Table~\ref{tab:DA_denoising}. The same architectures are used for Source only and Wasserstein.

\begin{table*}
\caption{Architectures and hyper-parameters: Domain adaptation for image denoising experiment}
\label{tab:DA_denoising}
\centering
\begin{tabular}{|c|c|}
\hline
Encoder & Decoder \\
\hline
\hline
Conv($3\to64$, $3\times 3$ kernel)  & Linear($2\to 128$) \\
+ReLU + MaxPool(2) &  Conv($128\to 64$, $3\times 3$ kernel) \\
Conv($64\to128$, $3\times 3$ kernel)  & + ReLU + Upsample(2) \\
+ReLU + MaxPool(2) & Conv($64\to 64$, $4\times 4$ kernel) \\
Conv($128\to128$, $3\times 3$ kernel)  & + ReLU + Upsample(4) \\
+ReLU + MaxPool(2) & Conv($64 \to 3$, $3\times 3$ kernel) \\
Conv($128\to128$, $3\times 3$ kernel)  & \\
Linear($128\to 2$) & \\
\hline
\multicolumn{2}{|c|}{Domain discriminator}  \\
\hline
\hline
\multicolumn{2}{|c|}{Linear($2 \to 64$) + ReLU}  \\
\multicolumn{2}{|c|}{Linear($64 \to 64$) + ReLU}  \\
\multicolumn{2}{|c|}{Linear($64 \to 1$)}  \\

\hline
Generator (RW) & Discriminator (RW) \\
\hline
\hline
 Linear($2\to 128$) & Linear($2\to 128$) + ReLU \\
 Linear($128\to 128$) & Linear($128\to 128$) + ReLU \\
 Linear($128\to 2$) & Linear($128\to 1$) \\
\hline
\multicolumn{2}{|c|}{Hyperparameters} \\
\hline
\hline
Encoder learning rate &  $0.0002$   \\
Decoder learning rate &  $0.0002$ \\
Domain Discriminator learning rate & $0.0002$ \\
Generator (RW) learning rate & $0.0002$ \\
Discriminator (RW) learning rate & $0.0002$ \\
$\pi$ learning rate & $0.0005$ \\
Batch size & $128$ \\
Optimizer & Adam \\
Number of critic iters & $5$ \\
Initial iters & $5000$ \\
Weight clipping value & $[-0.01, 0.01]$\\
$\lambda$ & $1$ \\
\hline
\end{tabular}
\end{table*}

\subsection{Adversarial clustering}

For adversarial clustering in imbalanced MNIST dataset (Section 5 of the main paper), the architectures and hyper-parameters used are given in Table~\ref{tab:MNIST_mixture}.

\begin{table*}
\caption{Architectures and hyper-parameters: Mixture models on imbalanced-MNIST3 dataset}
\label{tab:MNIST_mixture}
\centering

\begin{tabular}{|c|c|}
\hline
Generator & Discriminator\\   
\hline
\hline
ConvTranspose($100\to 256$, $4\times 4$ kernel, stride 1)  & Spectralnorm(Conv($1\to 64$, $4\times 4$ kernel, stride 2)) \\
Batchnorm + ReLU & LeakyReLU(0.2) \\
ConvTranspose($256\to 128$, $4\times 4$ kernel, stride 2) & Spectralnorm(Conv($64\to 128$, $4\times 4$ kernel, stride 2)) \\
Batchnorm + ReLU & LeakyReLU(0.2) \\
ConvTranspose($128\to 64$, $4\times 4$ kernel, stride 2) & Spectralnorm(Conv($128\to 256$, $4\times 4$ kernel, stride 2)) \\
Batchnorm + ReLU & LeakyReLU(0.2)\\
ConvTranspose($64\to 1$, $4\times 4$ kernel, stride 2) & Spectralnorm(Conv($256\to 1$, $4\times 4$ kernel, stride 1)) \\
Tanh() & \\
\hline
\multicolumn{2}{|c|}{Hyperparameters} \\
\hline
\hline
Discriminator learning rate & $0.00005$ \\
Generator learning rate & $0.0001$ \\
$\pi$ learning rate & $0.001$ \\
Batch size & $64$ \\
Optimizer & RMSProp \\
Number of critic iters & $5$ \\
Weight clip & $[-0.01, 0.01]$ \\
$\lambda_{reg}$ & $0.01$ \\
\hline
\end{tabular}
\end{table*}

\subsection{Hypothesis testing}

For hypothesis testing experiment (Section 6 of the main paper), the same model architectures and hyper-parameters as the MOG experiment (Table~\ref{tab:MOG_relaxed_WGAN}) was used.

\end{appendix}

\end{document}